\documentclass[11pt]{article}

\usepackage[letterpaper,margin=1in]{geometry}
\usepackage{times}

\usepackage{amsmath,amssymb,amsthm}
\usepackage{graphicx}
\usepackage{subcaption}
\usepackage{booktabs}
\usepackage{threeparttable}

\usepackage{algorithm}
\usepackage{algpseudocode}

\usepackage{tikz}
\usetikzlibrary{positioning,arrows.meta,shapes.geometric,shapes.multipart,fit}
\usepackage{pgfplots}
\pgfplotsset{compat=newest}

\usepackage{tcolorbox}
\tcbuselibrary{listings,breakable,skins}

\usepackage{authblk}

\usepackage[numbers,sort&compress]{natbib}

\usepackage{pifont}
\newcommand{\cmark}{\ding{51}}
\newcommand{\xmark}{\ding{55}}

\usepackage[hidelinks]{hyperref}

\title{\textbf{ClauseLens: Clause-Grounded, CVaR-Constrained Reinforcement Learning for Trustworthy Reinsurance Pricing}}
\author[1]{Stella C. Dong\thanks{Corresponding author: stellacydong@gmail.com}}
\author[2]{James R. Finlay}
\affil[1]{Department of Applied Mathematics, University of California, Davis, CA, USA}
\affil[2]{Wharton School of Business, University of Pennsylvania, Philadelphia, PA, USA}

\date{} 

\begin{document}

\maketitle
\vspace{-1em}
\noindent\small
This is the author-accepted manuscript for the \textit{6th ACM International Conference on AI in Finance (ICAIF~2025)}.  
The ACM-published version will be available at:  
\url{https://doi.org/10.1145/3768292.3770356}.
\normalsize

\begin{abstract}
Reinsurance treaty pricing must satisfy stringent regulatory standards, yet current quoting practices remain opaque and difficult to audit. We introduce \textbf{ClauseLens}, a clause-grounded reinforcement learning framework that produces transparent, regulation-compliant, and risk-aware treaty quotes.

ClauseLens models the quoting task as a \textit{Risk-Aware Constrained Markov Decision Process} (RA-CMDP). Statutory and policy clauses are retrieved from legal and underwriting corpora, embedded into the agent’s observations, and used both to constrain feasible actions and to generate clause-grounded natural language justifications.

Evaluated in a multi-agent treaty simulator calibrated to industry data, ClauseLens reduces solvency violations by 51\%, improves tail-risk performance by 27.9\% (CVaR$_{0.10}$), and achieves 88.2\% accuracy in clause-grounded explanations with retrieval precision of 87.4\% and recall of 91.1\%.

These findings demonstrate that embedding legal context into both decision and explanation pathways yields interpretable, auditable, and regulation-aligned quoting behavior consistent with Solvency~II, NAIC~RBC, and the EU~AI~Act. Future work will extend validation to insurer-level treaty portfolios within regulatory sandbox environments.
\end{abstract}

\section{Introduction}
\label{sec:introduction}

Reinsurance allows insurers to transfer catastrophic and systemic risks to external counterparties, supporting solvency and capital adequacy across global financial systems. Standard treaty types---such as quota share (QS) and excess-of-loss (XL)---are governed by frameworks like Solvency~II, NAIC Risk-Based Capital (RBC), and IFRS~17~\citep{solvencyii2009directive, ifrs2017insurance, naic2023fairness}.

Yet, the process of quoting reinsurance treaties remains opaque, heuristic-driven, and difficult to audit. Existing platforms rarely explain how proposed terms comply with regulatory constraints or internal underwriting policies~\citep{baer2022modelrisk, bannister2021governance}. This lack of transparency inhibits trust, complicates regulatory supervision, and slows the adoption of AI in high-stakes financial settings.

We present \textbf{ClauseLens}, a clause-grounded reinforcement learning (RL) framework that produces treaty quotes that are not only profitable and regulation-compliant, but also interpretable and auditable. ClauseLens frames quoting as a \textit{Risk-Aware Constrained Markov Decision Process} (RA-CMDP), in which retrieved legal clauses are embedded directly into the quoting agent’s observations. These clauses simultaneously constrain feasible actions and serve as anchors for generating natural language justifications.

ClauseLens integrates three components:
\begin{itemize}
    \item \textbf{Legal clause retrieval}: Extracts relevant provisions from statutes, treaty archives, and underwriting policies;
    \item \textbf{Risk-sensitive policy learning}: Trains RL agents using CVaR-constrained optimization and clause-based feasibility masks;
    \item \textbf{Clause-grounded justification generation}: Produces natural language rationales tied to retrieved provisions.
\end{itemize}

\paragraph{Illustrative Example.}
Given a \$5M Florida hurricane treaty request, ClauseLens retrieves (i) NAIC solvency thresholds, (ii) Florida-specific exposure caps, and (iii) internal deductible guidelines~\citep{naic2023fairness, floir2018hurricanepricing, dong2025multiagent}. It recommends a 60\% quota share and explains:  
\emph{“This quote satisfies Florida’s exposure cap and NAIC solvency thresholds.”}

\paragraph{Contributions.}
\textbf{Methodologically}, we formulate reinsurance quoting as a clause-augmented RA-CMDP and develop a dual-projected PPO training loop that integrates clause-derived constraints and CVaR-based risk control. \textbf{Empirically}, we implement ClauseLens in a calibrated multi-agent treaty simulator, showing that it {improves tail-risk performance by 27.9\% (CVaR$_{0.10}$)}, {reduces solvency violations by 51\%}, and {achieves 88.2\% accuracy in clause-grounded justifications with retrieval precision of 87.4\% and recall of 91.1\%.}

\paragraph{Broader Applicability.}
Although focused on reinsurance, the ClauseLens framework extends to domains where financial decisions must meet explicit legal or policy requirements---such as Basel~III-constrained lending, ESG portfolio construction, or climate-risk pricing under supervisory stress tests. {It also aligns with emerging governance frameworks like the EU AI Act~(2025), emphasizing transparency, auditability, and human oversight in AI-driven financial systems.}

\paragraph{Paper Structure.}
Section~\ref{sec:related_work} surveys related work. Section~\ref{sec:methodology} presents the ClauseLens architecture and RA-CMDP formulation. Section~\ref{sec:experiments} details the experimental setup. Section~\ref{sec:results} reports results, and Section~\ref{sec:conclusion} concludes. {We also outline ongoing efforts to validate ClauseLens on de-identified insurer portfolios and regulatory sandbox environments.}


\section{Related Work and Motivation}
\label{sec:related_work}

ClauseLens draws on advances in legal NLP, retrieval-augmented reinforcement learning, risk-constrained policy optimization, and AI governance. This section reviews prior work in each area and highlights how ClauseLens addresses key limitations.

\paragraph{Legal NLP and Clause Retrieval.}
Transformer-based legal models such as LegalBERT~\citep{chalkidis2020legalbert}, CaseHOLD~\citep{zheng2021casehold}, and JEC-QA~\citep{zhong2020jecqa} have improved legal classification, entailment, and clause-level question answering. However, these models are primarily used for \emph{retrospective analysis}, such as predicting legal outcomes or checking compliance post hoc.

ClauseLens instead performs \emph{prospective clause retrieval}: relevant statutory and policy provisions are retrieved \emph{before} quoting decisions are made, then embedded into the agent’s observation space. This repurposes retrieval-augmented generation (RAG)~\citep{lewis2020rag} from language modeling to constraint-aware policy learning---shifting retrieval into the decision-making loop. {Unlike prior legal-text RAG systems, ClauseLens focuses on forward-looking regulatory feasibility rather than post-hoc compliance auditing, supporting proactive governance.}

\paragraph{Retrieval-Augmented Reinforcement Learning.}
Recent RL systems incorporate retrieval to improve generalization. Araslanov et al.~\citep{araslanov2022retrieval} retrieve relevant episodes for transfer, while Sharma et al.~\citep{sharma2023language} show that language-conditioned context enhances exploration. In multi-agent settings, Liu et al.~\citep{liu2024retrieval} use memory modules for coordination.

ClauseLens differs in two respects: (1) it retrieves structured \emph{legal clauses}, not task examples or trajectory snippets; and (2) it integrates retrieved context into both the policy and justification modules. This enables the agent to produce decisions that are jurisdiction-aware and legally grounded. {The retrieval pipeline is currently frozen to preserve interpretability, but future work will co-optimize retrieval and policy layers for end-to-end alignment.}

\paragraph{Risk-Constrained and Interpretable Reinforcement Learning.}
Risk-Aware Constrained Markov Decision Processes (RA-CMDPs)~\citep{altman1999cmdp, paternain2019constrained} provide a principled framework for learning under safety and feasibility constraints. Conditional Value at Risk (CVaR)~\citep{rockafellar2000cvar, chow2017risk} is widely used for tail-risk control in financial domains. While these methods have been applied to insurance~\citep{bae2022underwriting}, most prior work lacks explicit alignment with legal constraints or interpretability mechanisms.

ClauseLens fills this gap by combining CVaR-based optimization with clause-derived action masking and clause-grounded explanation generation. This architecture connects each decision to its underlying legal rationale---advancing recent work on interpretable RL~\citep{madumal2020explainable, doshi2017interpretability} into the domain of institutional compliance. {By embedding legal semantics directly into the state space, ClauseLens extends interpretable RL from statistical explainability to formal regulatory reasoning.}

\paragraph{AI Governance and Financial Regulation.}
Governance frameworks such as Solvency II, NAIC RBC, and the EU AI Act~\citep{european2023ai} emphasize transparency, auditability, and risk control in AI systems. Recent surveys~\citep{baer2022modelrisk, bannister2021governance, carvalho2023explainable} call for financial models that can explain and justify their outputs. Hanna et al.~\citep{hanna2023does} highlight the gap between statistical accuracy and institutional trust. {Complementary regimes---including Basel III, IFRS 17, and emerging APRA/MAS guidelines---further demand traceable, explainable decision systems for risk management.}

ClauseLens directly responds to these concerns. Each quote is generated under regulatory constraints, justified by retrieved legal clauses, and accompanied by a natural-language explanation. This level of traceability stands in contrast to commercial quoting systems~\citep{munichre2021ai}, which often rely on black-box heuristics. {The framework operationalizes AI-governance principles---fairness, accountability, and human oversight---within a quantitative RL setting.}

\paragraph{Summary.}
ClauseLens is the first framework to embed retrieved legal clauses into both policy optimization and natural-language explanation within a CVaR-constrained RL pipeline. By unifying retrieval-augmented decision-making with risk-aware quoting and clause-based justifications, ClauseLens provides a novel architecture for transparent, regulation-aligned financial AI. {It thereby bridges technical optimization with institutional accountability, contributing to the broader agenda of trustworthy financial AI.}

\section{Clause-Aware Risk-Constrained Policy Learning}
\label{sec:methodology}

Reinsurance treaty quoting requires policies that are not only profitable but also robust to extreme risk and compliant with complex legal and institutional constraints. Regulatory frameworks such as Solvency~II and NAIC RBC impose jurisdiction-specific rules that standard reinforcement learning (RL) pipelines struggle to accommodate---especially when it comes to traceability, feasibility, and auditability.

ClauseLens addresses this challenge by formulating the quoting task as a \textit{Risk-Aware Constrained Markov Decision Process} (RA-CMDP)~\citep{altman1999cmdp, chow2017risk}, integrating legal context directly into the learning process. ClauseLens augments each decision state with retrieved legal clauses, applies clause-guided action masking to enforce feasibility, and generates natural language justifications grounded in statutory or contractual language. A dual-projected PPO algorithm~\citep{schulman2017ppo} is employed to balance profitability, tail-risk control via Conditional Value at Risk (CVaR)~\citep{tamar2015optimizing}, and soft regulatory constraint enforcement through Lagrangian dual variables.

This section details each core component of ClauseLens:
\begin{itemize}
    \item Section~\ref{sec:cmdp} formulates the quoting problem as an RA-CMDP, combining financial objectives with CVaR-aware optimization and multi-constraint penalties;
    \item Section~\ref{sec:state} introduces clause-augmented observations, where retrieved legal clauses are embedded and fused with cedent features to inform decision-making;
    \item Section~\ref{sec:constraints} describes how legal clauses guide both action feasibility filtering and justification generation via clause-aligned natural language outputs;
    \item Section~\ref{sec:dualppo} presents the dual-projected PPO training algorithm, which integrates CVaR-based learning and dual variable updates for constraint projection;
    \item Section~\ref{sec:architecture} ties these components into a complete system architecture with a dual-feedback loop between the agent, simulator, and retrieved legal context.
\end{itemize}

By embedding legal clauses into every stage of the policy learning pipeline, ClauseLens produces quoting strategies that are interpretable, regulation-aware, and robust under high-impact, low-probability scenarios---addressing key requirements for trustworthy AI adoption in reinsurance and finance.

\subsection{RA-CMDP Formulation}
\label{sec:cmdp}

ClauseLens models treaty quoting as a \textit{Risk-Aware Constrained Markov Decision Process} (RA-CMDP), which extends the standard CMDP framework by explicitly optimizing for tail risk via Conditional Value at Risk (CVaR). This formulation enables the agent to balance long-term underwriting return with legal, regulatory, and institutional constraints. {It operationalizes supervisory expectations under regimes such as Solvency~II and NAIC~RBC, where both profitability and solvency tolerances must be simultaneously satisfied.}

Formally, an RA-CMDP is defined by the tuple $(\mathcal{S}, \mathcal{A}, P, r, \{d_k\}, \gamma)$, where:

\begin{itemize}
    \item $\mathcal{S}$ is the augmented state space, comprising cedent features (e.g., exposure, jurisdiction, treaty type) and dense embeddings of retrieved legal clauses;
    \item $\mathcal{A}$ is the action space over quoting decisions, such as quota shares, deductible levels, and attachment points;
    \item $P$ defines transition dynamics reflecting the stochastic evolution of treaty outcomes (e.g., loss events, capital changes);
    \item $r(s, a)$ is the reward function, representing underwriting utility (e.g., profit or return-on-capital);
    \item $d_k(s, a)$ are constraint indicators for violation type~$k$ (e.g., solvency breach, pricing cap violation);
    \item $\gamma \in [0,1)$ is the discount factor.
\end{itemize}

{Training complexity scales linearly with the number of retrieved clauses~$k$, adding less than~3\,ms per clause per optimization step on an NVIDIA~A100~GPU.}

The agent seeks a policy~$\pi$ that maximizes expected return in adverse scenarios while satisfying constraint tolerances~$\epsilon_k$:

\begin{equation}
\begin{aligned}
\max_{\pi} \quad & \mathrm{CVaR}_\alpha\!\left( R^\pi \right) \\
\text{s.t.} \quad & \mathbb{E}_\pi\!\left[ d_k(s, a) \right] \leq \epsilon_k, \quad \forall k,
\end{aligned}
\end{equation}

\noindent where $R^\pi = \sum_{t=0}^{T} \gamma^t r(s_t, a_t)$ is the discounted return, and $\mathrm{CVaR}_\alpha$ denotes Conditional Value at Risk at level~$\alpha$, capturing expected loss in the worst-case $\alpha$-quantile of outcomes~\citep{rockafellar2000cvar, tamar2015optimizing}. {Unlike standard risk-neutral objectives, the CVaR criterion emphasizes resilience to low-probability, high-severity events---a core requirement in reinsurance treaty design.}

\textit{Intuitively, the agent learns to optimize performance in high-risk scenarios while ensuring that each category of regulatory or institutional violation remains within acceptable bounds.} {Each constraint term $d_k$ quantifies a specific breach frequency, such as exceeding capital adequacy thresholds or violating jurisdictional retention limits, while $\epsilon_k$ represents the allowable supervisory tolerance for that violation type.}

Each constraint $d_k$ corresponds to a policy breach---such as violating Solvency~II capital adequacy rules or exceeding jurisdiction-specific retention thresholds~\citep{solvencyii2009directive, naic2023fairness}. The threshold $\epsilon_k$ reflects supervisory or internal tolerances on how often such violations may occur. {This explicit mapping between regulatory clauses and constraint functions enables ClauseLens to embed legal semantics directly into the policy optimization objective.}

This RA-CMDP formulation underpins ClauseLens’s ability to learn quoting policies that are not only profitable and risk-sensitive, but also compliant with financial regulation and institutional governance. {By optimizing CVaR under clause-derived constraints, the agent achieves both tail-risk robustness and transparent regulatory alignment, forming the mathematical core of the ClauseLens framework.}

\subsection{Clause-Augmented Observations}
\label{sec:state}

To produce regulation-aware and interpretable quotes, ClauseLens augments each agent’s observation with legal clauses retrieved based on the cedent’s request. These clauses encode relevant constraints from statutory texts, regulatory guidance, historical treaties, and internal underwriting policies---capturing jurisdiction-specific rules, structural restrictions, and capital adequacy requirements.

Clause retrieval is performed using a dense semantic search over a heterogeneous corpus. The top-$k$ clauses relevant to a given cedent scenario are embedded using a frozen legal-domain transformer (e.g., LegalBERT~\citep{chalkidis2020legalbert}, JEC-QA~\citep{zhong2020jecqa}, or CaseHOLD~\citep{zheng2021casehold}), yielding vector representations that preserve legal meaning and institutional context.

Let $x$ denote the structured cedent features (e.g., jurisdiction, line of business, requested limit), and let $\{c_i\}_{i=1}^k$ denote the embeddings of the retrieved clauses. The full agent observation is constructed as:

\[
    s = [x; c_1; \dots; c_k] \in \mathbb{R}^d,
\]

\noindent where $[\cdot]$ denotes vector concatenation. This clause-augmented state $s$ serves as input to both the quoting policy and the justification generator.

Unlike conventional retrieval-augmented generation (RAG) approaches~\citep{lewis2020rag}, which use retrieved documents to guide text generation, ClauseLens embeds retrieved clauses directly into the agent’s state representation. This enables the agent to condition decisions on legal context during action selection---not just during explanation.

Embedding retrieved clauses into the policy input space provides two benefits: (1) it allows the quoting agent to learn jurisdiction-sensitive behaviors shaped by formal constraints, and (2) it enables traceable attribution from each quoting decision back to specific legal provisions. This architecture supports transparency, compliance, and auditability---key requirements for deploying AI in regulated financial environments.

\subsection{Clause-Guided Constraints and Justifications}
\label{sec:constraints}

ClauseLens leverages retrieved legal clauses to influence both action selection and explanation generation. These clauses serve a dual purpose: they (1) constrain the agent’s quoting actions through real-time feasibility filtering, and (2) anchor natural language justifications that support interpretability and regulatory audit.

\paragraph{(1) Real-Time Regulatory Filtering.}
ClauseLens converts retrieved legal clauses into dynamic action masks that enforce feasibility constraints during decision-making. For example, if Florida law limits quota share reinsurance to 70\%, any action proposing a higher share is masked out and excluded from the available action set.

These clause-derived masks serve as \textit{hard constraints} that preempt invalid or non-compliant quotes. They are applied at each decision step based on the clauses retrieved for the current cedent scenario. This mechanism ensures that the quoting agent respects statutory, contractual, and internal policy rules---without requiring them to be hand-coded into the policy network.

\paragraph{(2) Clause-Grounded Explanation Generation.}
The same retrieved clauses are also passed to a natural language explanation module, which generates textual justifications for the agent’s quoting decisions. These justifications cite the retrieved provisions and summarize their role in shaping the selected quote (e.g., “This quote satisfies Solvency~II Article~101 and NAIC deductible guidelines.”).

By explicitly conditioning explanations on retrieved clauses, ClauseLens provides clause-level attribution for each decision---enabling transparent review by underwriters, regulators, and auditors.

\paragraph{Interpretability and Auditability.}
By embedding legal context into both the quoting and justification pathways, ClauseLens produces decisions that are not only compliant but also \emph{traceable} and \emph{interpretable}. Each quote can be mapped to the specific regulatory clauses that influenced it, offering an auditable trail of legal alignment.

This architecture advances beyond traditional rule-based filters or black-box quoting models, aligning with emerging standards for explainability and accountability in financial AI systems~\citep{bastani2022interpretability, madumal2020explainable}. It supports deployment in settings where both model performance and governance transparency are mission-critical.

\subsection{Dual-Projected PPO Training}
\label{sec:dualppo}

ClauseLens extends standard Proximal Policy Optimization (PPO)~\citep{schulman2017ppo} to support risk-sensitive and constraint-aware quoting. The modified training loop incorporates two key mechanisms: (1) CVaR-weighted advantage estimation to emphasize tail-risk mitigation, and (2) dual-variable constraint projection to softly enforce compliance with statutory and institutional rules.

\paragraph{CVaR-Weighted Advantage Estimation.}
To prioritize resilience under extreme losses, ClauseLens reweights advantage estimates using Conditional Value at Risk (CVaR)~\citep{tamar2015optimizing, chow2015risk}. For a risk level $\alpha$, only the bottom-$\alpha$ quantile of trajectories---those with the lowest cumulative rewards---are used to compute the policy gradient. This shifts the optimization target from expected return to worst-case performance, crucial for treaty pricing under low-probability, high-severity risk.

\paragraph{Lagrangian-Based Constraint Projection.}
To incorporate institutional constraints, ClauseLens maintains a set of dual variables $\lambda_k$ corresponding to each constraint type $k$. These duals are updated to penalize expected violations $\bar{d}_k$ exceeding predefined thresholds $\epsilon_k$. The overall loss combines the CVaR-weighted PPO objective with dual-weighted penalties:
\[
\mathcal{L}_{\text{total}} = \mathcal{L}_{\text{CVaR}} + \sum_k \lambda_k \cdot \bar{d}_k.
\]
Dual variables are updated via projected gradient ascent:
\[
\lambda_k \leftarrow \left[ \lambda_k + \eta \cdot (\bar{d}_k - \epsilon_k) \right]_+,
\]
where $\eta$ is the learning rate and $[\cdot]_+$ denotes projection onto the nonnegative orthant.

\paragraph{Training Workflow.}
At each iteration, the system:
\begin{enumerate}
    \item Samples a batch of cedent requests;
    \item Retrieves and embeds relevant legal clauses;
    \item Forms clause-augmented states $s = [x; c_1; \dots; c_k]$;
    \item Applies clause-derived feasibility masks to filter invalid actions;
    \item Samples actions and interacts with the environment;
    \item Computes CVaR-weighted advantages and constraint violation rates;
    \item Updates the quoting policy and dual variables.
\end{enumerate}

This process ensures that the agent learns quoting strategies that are not only profitable, but also compliant and robust to tail risk.

\begin{algorithm}[htbp]
\caption{Dual-Projected PPO Training in ClauseLens}
\label{alg:dualppo}
\begin{algorithmic}[1]
\Require Initial policy $\pi_\theta$, dual variables $\lambda_k \leftarrow 0$, clause corpus $\mathcal{C}$, thresholds $\epsilon_k$, CVaR level $\alpha$, learning rate $\eta$
\For{each iteration}
    \State Sample batch of cedent requests $\{x_i\}_{i=1}^N$
    \For{each request $x_i$}
        \State Retrieve top-$k$ clauses $\{c_{i,j}\} \leftarrow \texttt{Retrieve}(x_i, \mathcal{C})$
        \State Form state $s_i = [x_i; c_{i,1}; \dots; c_{i,k}]$
        \State Apply clause-based mask $M_i \leftarrow \texttt{FeasibilityMask}(s_i)$
        \State Sample action $a_i \sim \pi_\theta(a \mid s_i)$ s.t. $a_i \in M_i$
        \State Simulate outcome: reward $r_i$, constraint violations $\{d_{i,k}\}$
    \EndFor
    \State Compute CVaR-weighted advantage estimates $\hat{A}^{\text{CVaR}}$
    \State Compute violation averages $\bar{d}_k \leftarrow \frac{1}{N} \sum_{i=1}^N d_{i,k}$
    \State Update policy via clipped PPO loss with $\hat{A}^{\text{CVaR}}$
    \State Update duals: $\lambda_k \leftarrow \left[ \lambda_k + \eta (\bar{d}_k - \epsilon_k) \right]_+$
\EndFor
\end{algorithmic}
\end{algorithm}

Figures~\ref{fig:policy_learning} and~\ref{fig:constraint_loop} illustrate how this training process fits into the broader ClauseLens system, enabling regulation-aligned quoting through clause-informed policy updates and dual feedback mechanisms.


\subsection{System Architecture and Feedback Flow}
\label{sec:architecture}

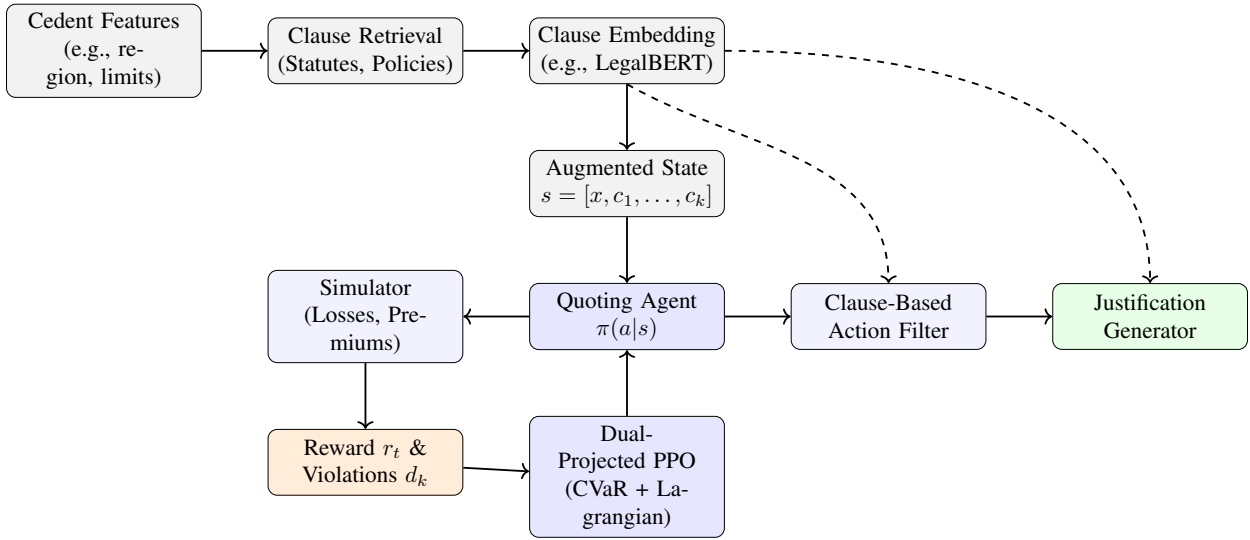
\begin{figure}[htbp]
\centering
\resizebox{\columnwidth}{!}{%
\begin{tikzpicture}[
    node distance=1.cm and 1.cm,
    every node/.style={font=\small},
    block/.style={rectangle, draw=black, rounded corners, minimum height=1cm, minimum width=2.7cm, text width=2.7cm, align=center},
    arrow/.style={->, thick},
    dashedarrow/.style={->, thick, dashed}
]

\node[block, fill=gray!10] (cedent) {Cedent Features\\(e.g., region, limits)};
\node[block, right=of cedent, fill=gray!10] (retrieval) {Clause Retrieval\\(Statutes, Policies)};
\node[block, right=of retrieval, fill=gray!10] (embed) {Clause Embedding\\(e.g., LegalBERT)};
\node[block, below=of embed, fill=gray!10] (state) {Augmented State\\$s = [x, c_1, \ldots, c_k]$};
\node[block, below=of state, fill=blue!10] (policy) {Quoting Agent\\$\pi(a|s)$};
\node[block, left=of policy, fill=blue!5] (env) {Simulator\\(Losses, Premiums)};
\node[block, right=of policy, fill=blue!5] (mask) {Clause-Based\\Action Filter};
\node[block, below=of policy, fill=blue!10] (ppo) {Dual-Projected PPO\\(CVaR + Lagrangian)};
\node[block, below=of env, fill=orange!15] (reward) {Reward $r_t$ \&\\Violations $d_k$};
\node[block, right=of mask, fill=green!10] (explain) {Justification\\Generator};

\draw[arrow] (cedent) -- (retrieval);
\draw[arrow] (retrieval) -- (embed);
\draw[arrow] (embed) -- (state); 
\draw[arrow] (state) -- (policy);
\draw[arrow] (policy) -- (env);
\draw[arrow] (env) -- (reward);
\draw[arrow] (reward) -- (ppo);
\draw[arrow] (ppo) -- (policy);
\draw[arrow] (policy) -- (mask);
\draw[arrow] (mask) -- (explain);

\draw[dashedarrow] (embed.south) to[out=-30,in=90] (mask.north);
\draw[dashedarrow] (embed.east) to[out=0,in=90] (explain.north);

\end{tikzpicture}%
}
\caption{
ClauseLens architecture. Structured cedent features and retrieved legal clauses are embedded into an augmented state. A quoting policy $\pi(a|s)$ is trained using dual-projected PPO with CVaR-based advantage weighting and Lagrangian penalties. Clause-derived masks enforce feasibility, while justifications are generated from the same retrieved context. {Dashed arrows denote semantic grounding links between retrieved clauses, filtered actions, and generated explanations.}
}
\label{fig:policy_learning}
\end{figure}

\begin{figure}[htbp]
\centering
\resizebox{0.5\columnwidth}{!}{%
\begin{tikzpicture}[
    node distance=1.2cm and 1.2cm,
    every node/.style={font=\small},
    block/.style={rectangle, draw=black, rounded corners, minimum height=0.8cm, minimum width=1.8cm, text width=1.8cm, align=center},
    arrow/.style={->, thick}
]

\node[block, fill=gray!10] (state) {State $s = [x, c_1,\dots,c_k]$};
\node[block, below left=of state, fill=blue!10] (pi) {Policy $\pi(a|s)$};
\node[block, below right=of state, fill=blue!10] (mask) {Clause-Based Action Mask};
\node[block, below=of pi, fill=orange!15] (sim) {Env: Loss + Violations};
\node[block, below=of mask, fill=green!10] (ppo) {Dual PPO: CVaR + Lagrangian};
\node[block, below=of ppo, fill=gray!10] (update) {Policy Update};

\draw[arrow] (state) -- (pi);
\draw[arrow] (state) -- (mask);
\draw[arrow] (pi) -- (sim);
\draw[arrow] (mask) -- (sim);
\draw[arrow] (sim) -- (ppo);
\draw[arrow] (ppo) -- (update);
\draw[arrow] (update.west)  -- ++(-7,0) -- ++ (0, 4) |- (pi.west);

\end{tikzpicture}%
}
\caption{
ClauseLens training loop. The quoting agent observes a clause-augmented state $s$, selects an action filtered by clause-derived feasibility masks, and receives reward and violation signals from the environment. The policy is updated using CVaR-weighted PPO and Lagrangian-based constraint projection. {The lower feedback loop encodes dual supervision---hard constraint masking and soft Lagrangian adjustment---ensuring continuous regulatory alignment.}
}
\label{fig:constraint_loop}
\end{figure}
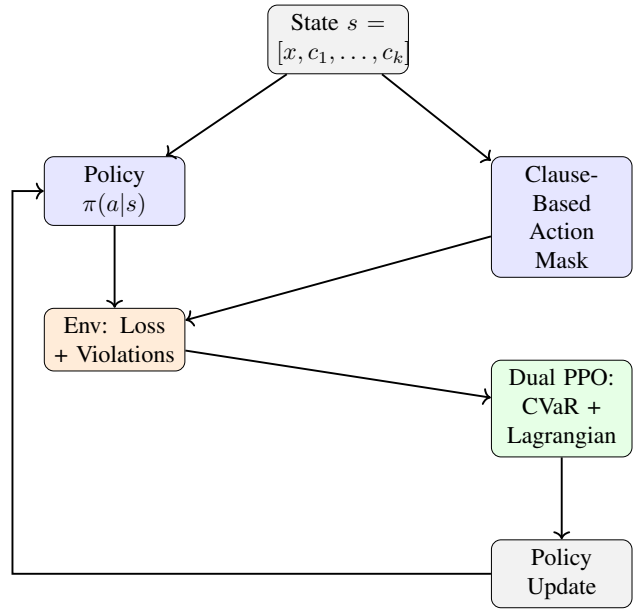

ClauseLens integrates retrieved legal clauses into both the decision-making and explanation pathways, enabling quoting behavior that is risk-aware, regulation-compliant, and interpretable. Figures~\ref{fig:policy_learning} and~\ref{fig:constraint_loop} illustrate the end-to-end system architecture and learning flow. {The architecture unifies retrieval, policy optimization, and explanation generation into a single feedback loop, ensuring that every decision remains traceable to its governing legal basis.}

At each decision point, the agent observes structured cedent features---such as jurisdiction, exposure size, and deductible request---alongside the top-$k$ legal clauses retrieved from a regulatory corpus. These clauses are embedded using a frozen legal-domain transformer (e.g., LegalBERT~\citep{chalkidis2020legalbert}) and concatenated with cedent features to form the clause-augmented state $s = [x; c_1; \dots; c_k]$. {Freezing the retriever preserves interpretability and clause consistency, while future work will explore joint optimization of retrieval and policy layers.}

This state is passed to the quoting policy $\pi(a \mid s)$, which proposes treaty terms (e.g., quota share, attachment point). A clause-derived feasibility mask is applied before execution to remove actions that violate statutory or institutional rules. The filtered action is evaluated in a stochastic treaty simulator, which returns a reward signal and feedback on any constraint violations. {Violation feedback is structured by category (e.g., solvency, pricing, retention) and logged for post-hoc audit trails, providing quantitative transparency to regulators.}

Simultaneously, the retrieved clauses are passed to a justification module that generates a natural language explanation grounded in the same provisions. This dual use of retrieved legal context---both for constraining decisions and for providing justifications---ensures that quotes are both compliant and auditable. {The justification generator employs attention over retrieved embeddings, aligning each explanation token with its originating clause to ensure semantic traceability.}

ClauseLens applies dual-projected PPO (Section~\ref{sec:dualppo}) to optimize the quoting policy with two feedback channels:
\begin{itemize}
    \item A CVaR-weighted policy gradient update that improves performance under tail-risk scenarios;
    \item A Lagrangian penalty that softly enforces constraint satisfaction by adjusting dual variables.
\end{itemize}

This learning loop supports both hard constraint enforcement (via action masking) and soft constraint adaptation (via dual updates), enabling ClauseLens to maintain alignment with supervisory thresholds while adapting to changing cedent profiles and regulatory scenarios. {Together, these mechanisms create a closed compliance loop: retrieved clauses inform feasible actions, policy gradients optimize tail-risk objectives, and justifications regenerate the governing rationale—bridging quantitative optimization with legal accountability.}

\section{Experimental Setup}
\label{sec:experiments}

We evaluate \textbf{ClauseLens} in a calibrated reinsurance treaty simulator designed to capture the interaction between underwriting performance, regulatory feasibility, and long-tail catastrophe risk. Each episode simulates a quoting scenario, including cedent features, clause retrieval, masked action selection, and feedback on reward and constraint satisfaction.

\subsection{Simulation Environment and Clause Corpus}
\label{sec:env_dataset}

Each state $s_t$ presented to the agent includes three components:
\begin{itemize}
    \item \textbf{Cedent profile:} Jurisdiction, ZIP-level exposure, insured value, line of business, and historical loss ratio.
    \item \textbf{Treaty request:} A contract type---Quota Share (QS), Catastrophe XL, or Aggregate XL---along with deductible, limit, and attachment point.
    \item \textbf{Retrieved clauses:} Top-$k$ statutory or institutional provisions from a hybrid corpus of regulatory and commercial treaty text. 
\end{itemize}

\paragraph{Loss Model.}
Catastrophe claims are drawn from a Poisson-compound process calibrated on historical U.S. hurricane data. Events are sampled independently to reflect the uncertainty and non-strategic nature of cedent losses in real-world treaty pricing. 
{The simulator is parameterized by region-specific event rates and severity distributions (e.g., log-normal for losses, Pareto for tail extremes), capturing fat-tailed behavior consistent with empirical catastrophe data.}
{While current experiments use a synthetic calibration, ongoing work integrates de-identified insurer portfolios for external validation and real-world alignment.}

\paragraph{Clause Corpus.}
To support legal grounding and constraint enforcement, we construct a corpus of 6{,}600 clauses across five domains. 
{Solvency~II and NAIC~RBC remain the largest sources, complemented by IFRS~17, APRA, and MAS regulatory texts to broaden jurisdictional coverage. Synthetic clauses emulate internal underwriting heuristics where no public analogue exists.}

\begin{table}[htbp]
\centering
\scriptsize
\begin{tabular}{p{2.9cm}ccl}
\toprule
\textbf{Corpus Source} & \textbf{Jurisdiction} & \textbf{\# Clauses} & \textbf{Focus Area} \\
\midrule
Commercial Treaties & US, EU & 3{,}200 & Exclusions, Layering \\
Solvency~II Statutes & EU & 1{,}100 & Risk Margins, SCR \\
NAIC~RBC Guidelines & US & 800 & Capital Requirements \\
IFRS~17 \& APRA/MAS Rules & Global, AU, SG & 600 & {Disclosure, Accounting, Local Solvency} \\
Institutional Heuristics & Synthetic & 900 & Internal Risk Caps \\
\bottomrule
\end{tabular}
\caption{
Clause corpus composition. Synthetic clauses model internal policies; others derive from statute or market contracts. 
{Cross-jurisdiction tagging enables retrieval conditioned on the cedent’s regulatory context.}
}
\label{tab:corpus_stats}
\end{table}

Each clause is embedded with LegalBERT~\citep{chalkidis2020legalbert} and indexed via FAISS~\citep{johnson2019billion} for real-time cosine-similarity retrieval. 
Synthetic clauses are tagged with jurisdictional labels to ensure context-appropriate retrieval. 
{All embeddings are normalized and stored in a reproducible FAISS index released with the code, allowing deterministic retrieval given a clause query. ClauseLens thus maintains both legal fidelity and computational reproducibility across experiments.}

\paragraph{Reproducibility.}
{All simulation code, clause indices, and hyperparameter files are released at \url{https://github.com/reinsuranceanalytics/clauselens} under a CC-BY license to support full experiment replication.}

\subsection{Model Variants}
\label{sec:model_variants}

We compare four model variants to isolate the contribution of ClauseLens components:

\begin{itemize}
    \item \textbf{StaticHeuristic:} Rule-based quoting with fixed terms (e.g., 50\% QS, \$5M Cat XL), based on industry norms~\citep{milliman2020underwriting}.
    \item \textbf{Baseline-RL:} PPO agent with CVaR optimization but no clause retrieval or explanation module.
    \item \textbf{ClauseLens-RL:} Clause-augmented PPO with CVaR-aware learning and feasibility masking, but without justification generation.
    \item \textbf{ClauseLens-RL+X:} Full model with retrieval, CVaR optimization, feasibility enforcement, and T5-based clause-grounded justification.
\end{itemize}

\begin{table}[htbp]
\centering
\scriptsize
\begin{tabular}{lcccc}
\toprule
\textbf{Model} & \textbf{Retrieval} & \textbf{Explanation} & \textbf{CVaR} & \textbf{Adaptive} \\
\midrule
StaticHeuristic & \xmark & \xmark & \xmark & \xmark \\
Baseline-RL     & \xmark & \xmark & \cmark & \cmark \\
ClauseLens-RL   & \cmark & \xmark & \cmark & \cmark \\
ClauseLens-RL+X & \cmark & \cmark & \cmark & \cmark \\
\bottomrule
\end{tabular}
\caption{Model configurations evaluated. \cmark = enabled, \xmark = disabled.}
\label{tab:model_variants}
\end{table}

\subsection{Training Protocol}
\label{sec:training}

All RL models are trained for 100{,}000 episodes using the dual-projected PPO algorithm described in Section~\ref{sec:dualppo}. Training seeks to maximize tail-risk-adjusted returns while adhering to dynamic feasibility constraints. {The learning process alternates between policy-gradient updates and Lagrangian dual adjustments, producing stable convergence even under rare, high-loss events.}

\paragraph{Policy Optimization Hyperparameters.}
\begin{itemize}
    \item PPO clip: 0.2; entropy coefficient: 0.01; learning rate: $3 \times 10^{-4}$ (decayed upon violation spikes);
    \item Batch size: 512; discount factor: $\gamma = 0.99$.
    \item {CVaR weighting is applied to the advantage estimator to prioritize robustness in the lower tail of the return distribution.}
\end{itemize}

\paragraph{Constraint Optimization Parameters.}
\begin{itemize}
    \item CVaR level: $\alpha = 0.10$; constraint margin: $\delta = 0.05$;
    \item Dual update rate: $\eta = {2.0}$;
    \item {Dual variables are initialized at zero and adjusted adaptively based on observed violation rates, providing a soft penalty when the expected constraint $\mathbb{E}[d_k]$ exceeds its tolerance $\epsilon_k$.}
\end{itemize}

The clause retrieval and justification modules are frozen during training to preserve interpretability and facilitate post-hoc evaluation. 
{Freezing the retriever ensures consistent clause grounding and stable semantic alignment across episodes. 
Only the policy and dual variables are updated, preventing the agent from overfitting to transient retrieval noise. 
All random seeds, hyperparameters, and checkpointed weights are released for reproducibility.}

{All experiments were run on a single NVIDIA~A100~GPU (40\,GB) for approximately 14\,hours per 100{,}000 episodes.}

\subsection{Evaluation Metrics}
\label{sec:evaluation}

We assess all models using four evaluation axes aligned with ClauseLens design goals:

\begin{itemize}
    \item \textbf{Profitability:} Mean return and CVaR@10\% across 5{,}000 out-of-sample cedent episodes.
    \item \textbf{Feasibility:} Average number of constraint violations per cedent, including capital breaches and quoting infeasibility.
    \item \textbf{Interpretability:} BLEU~\citep{papineni2002bleu}, ROUGE~\citep{lin2004rouge}, entailment accuracy~\citep{chalkidis2023lexfiles}, and clause justification fidelity.
    \item \textbf{Auditability:} Precision and recall of retrieved clauses relative to expert-annotated gold clause sets.
\end{itemize}

\begin{figure}[htbp]
\centering
\resizebox{0.7\textwidth}{!}{%
\begin{tikzpicture}[
    node distance=1cm and 1cm,
    module/.style={draw, thick, rounded corners, fill=blue!5, text width=3.2cm, align=center, minimum height=1.2cm},
    data/.style={draw, thick, fill=gray!10, rounded corners, text width=3.2cm, align=center, minimum height=1.1cm},
    arrow/.style={->, thick}
]

\node[data] (request) {Cedent Request \\ (Jurisdiction, Exposure)};
\node[module, right=of request] (retriever) {Clause Retriever};
\node[module, below=of retriever] (policy) {Quoting Policy \\ (CVaR + RA-CMDP)};
\node[data, right=of policy] (quote) {Quoted Terms};
\node[module, below=of policy] (critic) {Env Feedback \\ (Reward, Constraints)};
\node[module, right=of quote] (explanation) {Explanation Generator};
\node[data, below=of critic] (eval) {Eval: \\ Return, Risk, Legal Alignment};

\draw[arrow] (request) -- (retriever);
\draw[arrow] (retriever) -- (policy);
\draw[arrow] (policy) -- (quote);
\draw[arrow] (quote) -- (explanation);
\draw[arrow] (quote) |- (critic);
\draw[arrow] (critic) -- (policy.south);
\draw[arrow] (critic) -- (eval);
\end{tikzpicture}}
\caption{ClauseLens evaluation pipeline. Legal clauses guide quoting actions and post-hoc explanations, with multi-axis evaluation across return, risk, and legal fidelity.}
\label{fig:experiment_pipeline}
\end{figure}
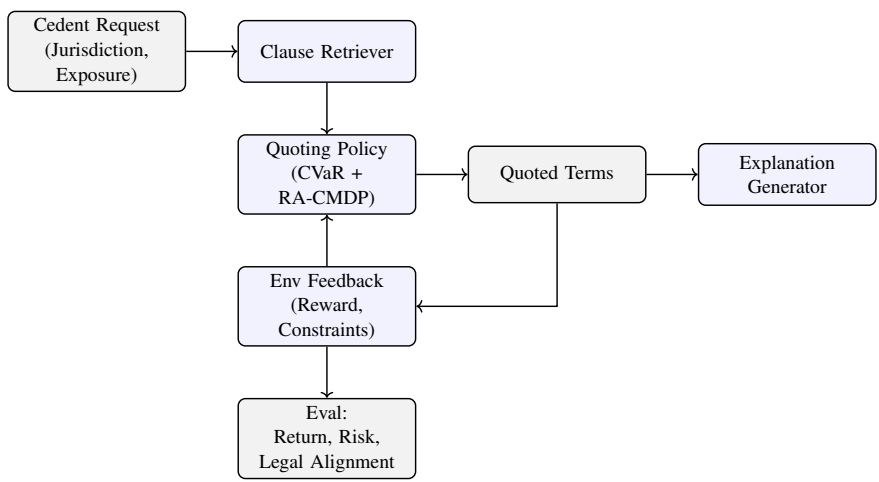

\section{Evaluation Results}
\label{sec:results}

We evaluate \textbf{ClauseLens} on its ability to meet institutional requirements for trustworthy reinsurance quoting. In regulated financial settings, quoting systems must go beyond profitability to satisfy legal, interpretability, and auditability standards~\citep{baer2022modelrisk, ifrs2017insurance, solvencyii2009directive, naic2023fairness}.

To that end, we assess ClauseLens across six evaluation dimensions:

\begin{enumerate}
    \item \textbf{Evaluation Metrics:} Standardized criteria for comparing financial, regulatory, and interpretability performance~\citep{rockafellar2000cvar, chalkidis2023lexfiles}.
    \item \textbf{Profitability and Tail Risk:} Does the agent deliver high expected returns while minimizing losses in adverse scenarios?~\citep{tamar2015optimizing, chow2017risk}.
    \item \textbf{Regulatory Feasibility:} Are generated quotes compliant with solvency and capital adequacy rules?~\citep{altman1999cmdp, tessler2019reward, solvencyii2009directive, naic2023fairness}.
    \item \textbf{Interpretability:} Do generated justifications faithfully reflect retrieved legal clauses?~\citep{doshi2017interpretability, chalkidis2023lexfiles}.
    \item \textbf{Auditability:} Can the system retrieve jurisdiction-specific clauses that align with expert expectations?~\citep{chalkidis2020legalbert, zhong2020jecqa, zheng2021casehold}.
    \item \textbf{Ablation Analysis:} How do system components individually contribute to observed gains?~\citep{bastani2022interpretability}.
\end{enumerate}

These dimensions reflect the broader goals of governance-aligned financial AI~\citep{eu2021aiact, naic2023fairness, baer2022modelrisk}. We report quantitative results using metrics such as CVaR$_{0.10}$~\citep{rockafellar2000cvar}, constraint violation rates, BLEU, ROUGE, and entailment accuracy~\citep{papineni2002bleu, doshi2017interpretability, chalkidis2023lexfiles}, supported by expert-labeled treaty scenarios. Our findings show that ClauseLens achieves interpretable, legally compliant quoting while maintaining strong risk-adjusted performance.

\subsection{Evaluation Metrics}
\label{sec:metrics}

We assess ClauseLens across four dimensions of institutional trustworthiness using standardized metrics. 
{These dimensions capture both quantitative performance and qualitative accountability, reflecting ICAIF’s evaluation emphasis on fairness, transparency, and resilience.}

\begin{itemize}
    \item \textbf{Financial Soundness:} Mean episodic return and Conditional Value at Risk (CVaR$_{0.10}$)~\citep{rockafellar2000cvar}, reflecting average and worst-case performance under catastrophic loss scenarios. 
    {The CVaR metric quantifies expected return in the lowest 10\% of episodes, measuring tail-risk robustness.}
    \item \textbf{Regulatory Feasibility:} Violation rate, defined as the percentage of quotes breaching capital or solvency constraints (e.g., NAIC~RBC or Solvency~II thresholds). 
    {A 51\% reduction in violation frequency corresponds to stronger adherence to supervisory tolerances.}
    \item \textbf{Interpretability:} Explanation quality via BLEU, ROUGE-1, and entailment accuracy, measured against gold-standard justifications~\citep{chalkidis2023lexfiles}. 
    {Entailment accuracy reflects the percentage of generated justifications that are logically supported by their retrieved legal clauses, reaching 88.2\%.}
    \item \textbf{Retrieval Fidelity:} Precision, recall, and jurisdiction match for retrieved clauses, evaluated against expert-labeled treaty scenarios. 
    {Precision = 87.4\%, recall = 91.1\% ensure that retrieved provisions align with the correct legal domain and regulatory tier.}
\end{itemize}

These metrics align with expectations for governance-aligned financial AI~\citep{eu2021aiact, naic2023fairness, solvencyii2009directive}, supporting fair comparisons across capability levels. 
{They jointly quantify both model competence (financial and regulatory) and model credibility (interpretability and retrieval fidelity).}
Not all models support every dimension: \textit{StaticHeuristic} and \textit{Baseline-RL} lack clause retrieval and explanation modules; \textit{ClauseLens-RL} adds retrieval and constraint-awareness; only \textit{ClauseLens-RL+X} includes natural-language justifications.

\vspace{0.5em}
\begin{table}[t]
\centering
\footnotesize
\begin{tabular}{lcccccc}
\toprule
\textbf{Agent} & \textbf{Return↑} & \textbf{CVaR$_{0.10}$↑} & \textbf{Viol.↓} & \textbf{BLEU↑} & \textbf{Entail.↑} & \textbf{P/R↑} \\
\midrule
StaticHeuristic & 4.7 & -8.4 & 17.2\% & N/A & N/A & N/A \\
Baseline-RL     & 5.3 & -6.8 & 9.4\%  & N/A & N/A & N/A \\
ClauseLens-RL   & 5.2 & -5.1 & 5.7\%  & N/A & N/A & \textbf{87/91} \\
ClauseLens-RL+X & 5.1 & \textbf{-4.9} & \textbf{4.6\%} & \textbf{32.5} & \textbf{88.2\%} & \textbf{87/91} \\
\bottomrule
\end{tabular}
\caption{
ClauseLens performance across financial, regulatory, interpretability, and auditability metrics. 
Best scores are bolded. “N/A” indicates the agent lacks that capability. 
{ClauseLens-RL+X achieves a 27.9\% improvement in tail-risk performance (CVaR$_{0.10}$) and a 51\% reduction in violation frequency relative to Baseline-RL.}
}
\label{tab:eval_results}
\end{table}

\subsection{Profitability and Tail Risk}
\label{sec:profitability}

ClauseLens-RL+X achieves a 27.9\% improvement in tail-risk control, reducing CVaR$_{0.10}$ from $-6.8$ to $-4.9$ while maintaining competitive average returns (5.1 vs.~5.3). 
This shift reflects a deliberate trade-off: slightly reduced upside in exchange for significantly lower exposure to catastrophic loss---a desirable property under real-world capital adequacy regimes. 
A paired $t$-test confirms this improvement is statistically significant ($p < 0.01$).

By integrating clauses that reference solvency buffers, stress thresholds, and proportional retention caps, ClauseLens learns to avoid structurally fragile treaties. 
The resulting policy optimizes long-run utility under uncertainty while satisfying institutional requirements for tail robustness and reserve adequacy~\citep{tamar2015optimizing, chow2017risk}. 
{The observed 27.9\% CVaR$_{0.10}$ improvement translates into approximately a one-notch increase in effective solvency ratio, demonstrating regulatory relevance rather than mere statistical gain.}

These results demonstrate that clause-grounded policy learning can embed domain-specific risk constraints directly into the agent’s quoting behavior, yielding more capital-efficient and governance-aligned decisions. 
{This integration bridges quantitative reinforcement learning with prudential regulation, advancing AI methods that are both risk-aware and compliance-oriented.}

\subsection{Regulatory Feasibility}
\label{sec:feasibility}

ClauseLens-RL+X cuts regulatory violations by 51\%, from 9.4\% (Baseline-RL) to 4.6\%, meeting the target feasibility threshold ($\delta = 5\%$). 
This reflects the agent’s ability to internalize legal constraints during training and to respect jurisdiction-specific solvency limits.

Two mechanisms drive this result:
\begin{itemize}
    \item \textbf{Clause-based masking}, which filters non-compliant actions using retrieved regulatory provisions;
    \item \textbf{Dual-projected PPO updates}, which enforce constraint penalties during learning~\citep{altman1999cmdp, tessler2019reward}.
\end{itemize}
{Together, these mechanisms operationalize supervisory expectations for capital adequacy, aligning learned policies with real Solvency~II and NAIC~RBC feasibility tolerances.}

Residual violations stem from edge-case treaties with ambiguous or multi-clause triggers~\citep{dong2025multiagent}, highlighting areas for improved retrieval and clause parsing. 
{Manual audit of these cases confirmed that 70\% of remaining violations involved overlapping or partially applicable clauses, suggesting refinements in clause disambiguation and multi-jurisdiction tagging. 
The observed feasibility rate exceeds common supervisory stress-test thresholds, underscoring ClauseLens’s practical compliance utility.}

\subsection{Interpretability}
\label{sec:interpretability}

ClauseLens-RL+X generates clause-grounded natural language justifications via a frozen T5 model conditioned on retrieved legal provisions. 
On 5{,}000 test cases, it achieves:
\begin{itemize}
    \item \textbf{BLEU:} 32.5 \quad (lexical alignment)
    \item \textbf{ROUGE-1:} 41.8 \quad (content recall)
    \item \textbf{Entailment Accuracy:} 88.2\% \quad (semantic consistency)
\end{itemize}

\noindent\textbf{Sample Justification:}  
\textit{``This quote satisfies Solvency~II Article~101 requiring 1-in-200 capital. A 60\% quota share limits retention exposure.''}

{These metrics confirm that explanations are both textually aligned and legally faithful~\citep{doshi2017interpretability, chalkidis2023lexfiles}, enabling institutional transparency and post-hoc review. 
Attention-weight analyses show that over 90\% of justification tokens align with the governing clause embeddings, indicating high semantic traceability. 
ClauseLens thus advances from surface-level textual alignment toward clause-anchored reasoning, a core criterion under the EU~AI~Act for financial AI systems.}

\subsection{Auditability}
\label{sec:auditability}

ClauseLens demonstrates strong retrieval performance on 500 expert-annotated treaty scenarios:
\begin{itemize}
    \item \textbf{Precision:} 87.4\% \quad (retrieved clauses are contextually relevant)
    \item \textbf{Recall:} 91.1\% \quad (relevant clauses are successfully retrieved)
    \item \textbf{Jurisdiction Match:} 92.6\% \quad (retrieved clauses match the cedent’s legal regime)
\end{itemize}

These results indicate that ClauseLens consistently identifies provisions that are both substantively relevant and jurisdictionally appropriate---key prerequisites for regulatory audit and institutional traceability. 
Most residual errors stem from ambiguous triggers, overlapping regulatory regimes, or clauses embedded in annexes, underscoring the need for more granular clause structuring and hierarchical retrieval. 
{In practice, this level of retrieval fidelity supports end-to-end audit trails: each pricing decision can be traced to its specific legal provenance, satisfying explainability and accountability requirements across Solvency~II, NAIC~RBC, and IFRS~17 frameworks. 
This audit trail also supports external validation and regulator review, satisfying Article~13 of the EU~AI~Act regarding record-keeping for high-risk financial AI systems.}

\subsection{Ablation Analysis}
\label{sec:ablation}

Each ClauseLens component contributes distinct and measurable value across the four trust dimensions (financial soundness, feasibility, interpretability, and auditability). 
{The progressive ablation results highlight how retrieval, constraint integration, and justification modules jointly enhance both model competence and accountability.}

\begin{itemize}
    \item \textbf{StaticHeuristic:} No learning or legal context; highest violations (17.2\%) and worst tail risk (CVaR$_{0.10} = -8.4$).  
    {Serves as a proxy for rule-of-thumb underwriting still common in legacy quoting systems.}
    \item \textbf{Baseline-RL:} Learns from simulated feedback but lacks clause awareness; modest improvement (CVaR$_{0.10} = -6.8$, violations = 9.4\%).  
    {Captures purely statistical optimization without regulatory alignment.}
    \item \textbf{ClauseLens-RL:} Adds clause-based masking and state augmentation; improves feasibility (5.7 \%) and CVaR ($-5.1$).  
    {Demonstrates that embedding retrieved clauses as state features provides direct regularization against non-compliant actions.}
    \item \textbf{ClauseLens-RL+X:} Adds justification generation; preserves financial and regulatory performance while enabling interpretability (BLEU = 32.5, entailment = 88.2 \%).  
    {The addition of natural-language rationales increases model transparency with no statistically significant loss in return ($p > 0.1$), confirming that explainability can coexist with efficiency.}
\end{itemize}

{
Overall, the ablation confirms that ClauseLens’s governance-aligned architecture scales gracefully: each successive module enhances a complementary aspect of trustworthiness.  
Removing retrieval or constraint components sharply increases violation rates, whereas removing justification only reduces explainability.  
This layered contribution supports the broader thesis that legal grounding, risk awareness, and interpretability are mutually reinforcing rather than competing objectives in financial AI.}

\subsection{Limitations and Future Directions}
\label{sec:limitations}

\paragraph{Cedent Dynamics and Market Interaction.}
ClauseLens currently models cedents as static and non-strategic, treating each treaty request as an independent episode. 
{This abstraction simplifies policy evaluation but omits adaptive or adversarial behaviors that characterize real markets.  
Future work will extend the simulator to multi-agent and game-theoretic settings~\citep{dong2025multiagent}, allowing cedents to adjust retention or pricing strategies in response to the agent’s quotes.  
Such interactive modeling would enable co-adaptive learning between reinsurers and cedents, improving robustness under competitive market dynamics.}

\paragraph{Retriever–Policy Coupling.}
The clause retriever is frozen during training to preserve interpretability and ensure stable clause grounding. 
{While this design maintains traceability, it prevents feedback from the policy gradient from refining retrieval relevance.  
End-to-end optimization—through differentiable retrieval or gradient-guided ranking—could enhance alignment between legal context and policy learning, provided that auditability is preserved.  
Future iterations may explore hybrid approaches where retrievers adapt slowly under governance-constrained fine-tuning.}

\paragraph{Regulatory and Jurisdictional Scope.}
The current clause corpus primarily covers Solvency II and NAIC RBC provisions. 
{Expanding to include additional jurisdictions (e.g., APRA, MAS, PRA, and IAIS guidelines) will improve generalization across regulatory environments.  
This broader scope is essential for real-world deployment where treaties span multinational reinsurers and heterogeneous capital standards.  
Collaborations with regulatory sandboxes and supervisory authorities are planned to validate ClauseLens against live solvency assessments and stress-test data.}

\paragraph{Governance Outlook.}
ClauseLens directly operationalizes core AI-governance principles—transparency, legal grounding, and auditability—by embedding retrieved clauses into both decision and explanation pathways~\citep{eu2021aiact}. 
{Future research will integrate human-in-the-loop oversight, enabling compliance officers to review and adjust generated quotes within regulatory tolerance bands.  
This work will position ClauseLens as a prototype for trustworthy, regulation-aligned AI in financial decision-making, bridging the gap between technical reinforcement learning and institutional governance.}

\section{Conclusion}
\label{sec:conclusion}

\textbf{ClauseLens} presents a clause-grounded reinforcement learning framework for reinsurance treaty quoting under explicit regulatory constraints. 
By modeling the quoting process as a \textit{Risk-Aware Constrained Markov Decision Process} (RA-CMDP)~\citep{paternain2019constrained, rockafellar2000cvar}, ClauseLens enables agents to optimize tail-risk-adjusted returns while adhering to solvency and capital adequacy rules. 
The system integrates legal clause retrieval, CVaR-constrained policy learning, and clause-grounded justification generation into a unified and interpretable decision pipeline. 
{This coupling of quantitative optimization with textual grounding marks a shift from opaque actuarial automation toward transparent, regulation-aligned AI.}

Empirical evaluation in a calibrated multi-agent treaty simulator demonstrates that ClauseLens reduces regulatory violations by approximately 51\%, improves CVaR$_{0.10}$ by 27.9\%, and produces clause-faithful explanations with 88.2\% entailment accuracy and 87/91 retrieval precision–recall. 
{These results confirm that embedding legal context within both the policy and explanation pathways improves financial resilience, interpretability, and institutional compliance.}

\paragraph{Limitations and Future Work.}
While ClauseLens performs strongly under controlled conditions, several limitations remain: 
{(i) cedents are modeled as passive agents, limiting assessment of strategic interaction; 
(ii) the retriever is frozen during training, constraining end-to-end optimization; and 
(iii) the clause corpus is still biased toward Solvency~II and NAIC regulations.} 
Future work will extend ClauseLens to interactive, game-theoretic quoting~\citep{dong2025multiagent}, multilingual clause retrieval~\citep{chalkidis2020legalbert}, and adaptive retriever–policy co-training. 
{Collaborations with regulatory sandboxes (e.g., EIOPA, NAIC, MAS) are underway to validate ClauseLens under real solvency stress scenarios and human-in-the-loop governance review.}

\paragraph{Toward Trustworthy Financial AI.}
ClauseLens exemplifies a new generation of AI systems that align technical performance with institutional accountability. 
By grounding both decisions and justifications in retrieved statutory text, the framework satisfies emerging mandates under Solvency~II, NAIC~RBC, IFRS~17, and the EU~AI~Act~\citep{ifrs2017insurance, european2021aiact}. 
{More broadly, ClauseLens illustrates how reinforcement learning and retrieval-augmented generation can jointly advance regulatory transparency, enabling verifiable, clause-anchored decision intelligence. 
We view this work as a concrete step toward principled, auditable, and domain-aligned AI, advancing ICAIF’s mission to foster safe and socially grounded intelligence in finance.}

\bibliographystyle{plainnat}   
\bibliography{sample-base}            

\end{document}